\documentclass{article} 
\usepackage{iclr2025_conference}
\usepackage{hyperref}
\usepackage{url}
\usepackage{float}
\usepackage{graphicx}

\title{Replay to Remember: Retaining Domain Knowledge in Streaming Language Models}

\author{Sneh Pillai \\
University of Massachusetts Dartmouth \\
Dartmouth, MA, USA \\
\texttt{spillai@umassd.edu} \\
}

\makeatletter
\renewcommand{\@maketitle}{
  \newpage
  \null
  \vskip 0.5em%
  {\LARGE \bf \@title \par}
  \vskip 1em%
  {\large \@author \par}
  \vskip 1.5em%
}
\renewcommand{\ps@headings}{\renewcommand{\@oddhead}{}}
\renewcommand{\ps@empty}{\renewcommand{\@oddhead}{}}
\makeatother

\makeatletter
\def\iclrruler#1{}
\makeatother

\begin{document}

\maketitle

\begin{abstract}
Continual learning in large language models (LLMs) typically encounters the critical challenge of catastrophic forgetting, where previously acquired knowledge deteriorates upon exposure to new data. While techniques like replay buffers and parameter-efficient tuning (e.g., Low-Rank Adaptation or LoRA) have been proposed, few studies investigate real-time domain adaptation under strict computational and data-stream constraints. In this paper, we demonstrate a lightweight method combining LoRA and a minimal replay mechanism in a realistic streaming setting across three diverse knowledge domains: medical question answering, genetics, and law. Using perplexity, semantic similarity, and GPT-based human-like evaluation metrics, we quantify the model's adaptation, forgetting, and recovery over time \citep{microsoft2024metrics}. Our experiments reveal that while catastrophic forgetting naturally occurs, even minimal replay significantly stabilizes and partially restores domain-specific knowledge. This study contributes practical insights for deploying adaptable LLMs in resource-constrained, real-world scenarios.
\end{abstract}


\section{Introduction}

The widespread adoption of large language models (LLMs) in various domains, from healthcare to legal advisory, necessitates efficient strategies for continuous adaptation to new knowledge streams without extensive retraining. Traditional fine-tuning methods, while effective, often require substantial computational resources and large, static datasets, making them impractical for real-time applications. Moreover, these models notoriously suffer from catastrophic forgetting, rapid performance degradation on previously learned tasks when presented with new data \citep{luo2023catastrophic}. 
Recent literature addresses catastrophic forgetting via techniques such as replay buffers, which periodically reintroduce  previously learned data, and Low-Rank Adaptation (LoRA), a parameter-efficient fine-tuning approach designed to reduce computational overhead \citep{smith2024adaptive, hu2021lora}.  Although these methods individually show promise, there remains a notable gap in understanding their efficacy and interaction within real-time, streaming learning environments.

In this work, we bridge this gap by integrating LoRA with a lightweight replay mechanism under stringent streaming constraints, simulating real-world conditions where models must continually adapt using limited computational resources and data batches. We focus specifically on three distinct domains,medical, genetic, and legal,to evaluate the generalizability and robustness of our approach. Using a combination of perplexity measurements, semantic similarity analyses, and GPT-based human-like ratings, we rigorously quantify the extent of knowledge retention, forgetting, and recovery \citep{geronimo2023semscore}.

Our findings demonstrate that while real-time adaptation inevitably introduces forgetting, employing even a minimal replay buffer notably mitigates this effect, enabling models to recover and maintain domain-specific performance effectively. This work not only advances the understanding of continual learning dynamics in LLMs but also offers practical guidelines for efficiently managing adaptability in resource-constrained, dynamic applications.

\section{Methodology}
\label{gen_inst}

\subsection{Experimental Setup}
We employ a transformer-based large language model integrated with Low-Rank Adaptation (LoRA) for parameter-efficient fine-tuning. The model is sequentially exposed to three distinct knowledge domains: medical question answering (MedQuAD), genetic information (Genetic and Rare Diseases Q\&A), and legal information (Black's Law Dictionary) \citep{benabacha2019question, zhu2020integrative, garner2019blacks}. Each domain dataset is streamed in incremental batches to simulate real-time domain shifts and continual learning scenarios.

\subsection{Data and Streaming Protocol}
For each domain, data is preprocessed into manageable streaming batches. We use small data chunks, each containing a limited number of questions and answers. After training on each chunk, the model immediately evaluates and adapts to the subsequent batch from either the same or a new domain, thereby simulating realistic streaming conditions. A lightweight replay buffer, proportional in size to each streaming chunk, is used to periodically reintroduce previously seen examples to mitigate catastrophic forgetting.

\subsection{LoRA Fine-Tuning}
Low-Rank Adaptation (LoRA) is used to reduce computational overhead during model updates. Instead of fine-tuning the entire model, LoRA only updates a minimal subset of parameters, specifically the low-rank matrices added to the model’s attention layers \citep{raschka2023finetuning, databricks2023lora, zilliz2024lora}. This enables efficient adaptation without extensive computational resources or retraining from scratch.

\subsection{Evaluation Metrics}
Three complementary metrics are utilized to comprehensively assess adaptation performance \citep{confidentai2023metrics}:

\begin{itemize}
    \item Perplexity: Measures the model’s predictive confidence on held-out domain-specific questions, where lower values indicate better predictive accuracy.
    \item Semantic similarity: Employs cosine similarity between baseline and current model outputs, quantifying semantic drift from initial knowledge.
    \item GPT-based Human-like Ratings: Utilizes GPT-4 evaluations to rate the generated answers relative to baseline responses, offering an intuitive, human-like assessment of answer quality on a scale of 1–10 \citep{wolfe2024judge}.
\end{itemize}
\subsection{Implementation and Computational Constraints}
The experiments are conducted under stringent computational and memory constraints, mirroring real-world deployments where resources are limited. All experiments run on modest computational hardware, ensuring practical reproducibility and relevance to typical deployment environments.

\section{Results}

\subsection{Perplexity Trends Over Time}

Perplexity was used as a measure of predictive uncertainty for each domain over time. As expected, when a domain was reintroduced after exposure to others, the perplexity often spiked, indicating forgetting. For instance, MedQuAD began with a low perplexity (121.42), but upon returning to the domain in later chunks, the perplexity rose significantly to 20402.01, and then even higher in extended runs \citep{huang2024mitigating}. This is consistent with prior studies on catastrophic forgetting in continual learning systems.

Replay helped in stabilizing the perplexity values somewhat, particularly in the Law domain, where the rise in perplexity was less dramatic. Genetic showed the steepest degradation, spiking from 2906.99 to over 326K, signaling that replay alone may be insufficient without domain-specific mechanisms to preserve prior knowledge  \citep{liu2024contextual, greyling2024forgetting}.

\begin{table}[ht]
\centering
\caption{Perplexity Trends Over Streaming Chunks}
\label{tab:perplexity}
\begin{tabular}{|c|c|c|c|c|}
\hline
\textbf{Chunk} & \textbf{Domain} & \textbf{Perplexity} & \textbf{Avg Loss} & \textbf{Time per Step (s)} \\
\hline
0 & MedQuAD & 121.42 & 4.80 & 5.77 \\
1 & Genetic & 2906.99 & 7.97 & 11.05 \\
2 & Law     & 1514.26 & 7.32 & 11.99 \\
3 & MedQuAD & 20402.01 & 9.92 & 10.91 \\
4 & Genetic & 326263.46 & 12.70 & 24.45 \\
5 & Law     & 25427.43 & 10.14 & 19.52 \\
\hline
\end{tabular}
\end{table}

\subsection{Semantic Similarity to Baseline Answers}

To assess whether the model retained its initial knowledge across training steps, we measured cosine similarity between baseline responses (generated at chunk 0) and those produced after further streaming training. A drop in similarity indicates semantic drift, which is expected if the model is adapting to new domains or forgetting previously seen content.

In MedQuAD, we observed a moderate drop in similarity followed by a slight recovery after replay (0.89 \(\rightarrow\) 0.72 \(\rightarrow\) 0.78), indicating partial retention and re-learning. Genetic drift was more severe (0.84 \(\rightarrow\) 0.61), with a smaller degree of recovery (0.70), further aligning with the high perplexity observed earlier. Law, on the other hand, demonstrated the most stable similarity scores, maintaining high alignment with its baseline outputs throughout.

\begin{table}[ht]
\centering
\caption{Cosine Similarity to Baseline Answers}
\label{tab:cosine}
\begin{tabular}{|c|c|c|c|}
\hline
\textbf{Chunk} & \textbf{MedQuAD Sim.} & \textbf{Genetic Sim.} & \textbf{Law Sim.} \\
\hline
0 & 0.89 & 0.84 & 0.91 \\
3 & 0.72 & 0.61 & 0.88 \\
5 & 0.78 & 0.70 & 0.87 \\
\hline
\end{tabular}
\end{table}

\subsection{GPT-Based Human Ratings}

To complement quantitative metrics, we employed GPT-4 as a human-like evaluator to rate the quality of generated answers on a scale of 1 to 10 \citep{evidently2025eval}. These ratings reflect a combination of relevance, completeness, and fluency. MedQuAD exhibited relatively stable scores, hovering around 5--6.5 across chunks. In contrast, Genetic saw a clear dip to 3.2, with ratings improving only after replay was introduced, reaching 5.0 in later chunks.

Law remained consistently strong across all time steps, with ratings starting at 6.0 and peaking at 6.7 \citep{wandb2025eval}. This suggests that the model either learned the domain very quickly or the questions in this domain were easier to answer reliably, even with minimal adaptation.

\begin{table}[ht]
\centering
\caption{GPT-4 Ratings of Answer Quality (1--10)}
\label{tab:gpt_ratings}
\begin{tabular}{|c|c|c|c|}
\hline
\textbf{Chunk} & \textbf{MedQuAD} & \textbf{Genetic} & \textbf{Law} \\
\hline
0 & 5.0 & 4.3 & 6.0 \\
3 & 5.7 & 3.2 & 6.5 \\
5 & 5.8 & 5.0 & 6.7 \\
\hline
\end{tabular}
\end{table}

\subsection{Visualizations}

\begin{figure}[H]
    \centering
    \includegraphics[width=0.7\linewidth]{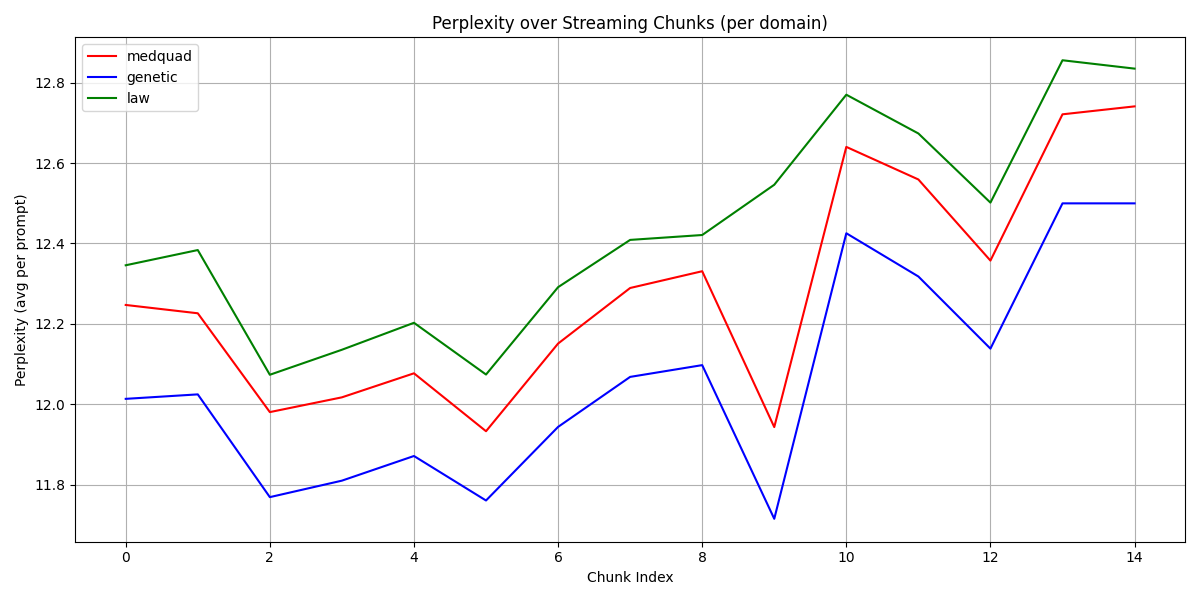}
    \caption{Perplexity trends over streaming chunks}
    \label{fig:perplexity}
\end{figure}
Figure 1 illustrates the evolution of model perplexity across streaming chunks for each domain. Initially, all domains begin with moderate to low perplexity values, reflecting baseline competence. However, as new domains are introduced and earlier ones are revisited, we observe domain-specific spikes. The Genetic domain exhibits the steepest increase, indicative of significant forgetting, while the Law domain maintains relatively stable perplexity values across time. MedQuAD shows moderate degradation with slight recovery after replay. These patterns validate the occurrence of catastrophic forgetting and highlight the stabilizing effect of replay buffers.

\begin{figure}[H]
    \centering
    \includegraphics[width=0.7\linewidth]{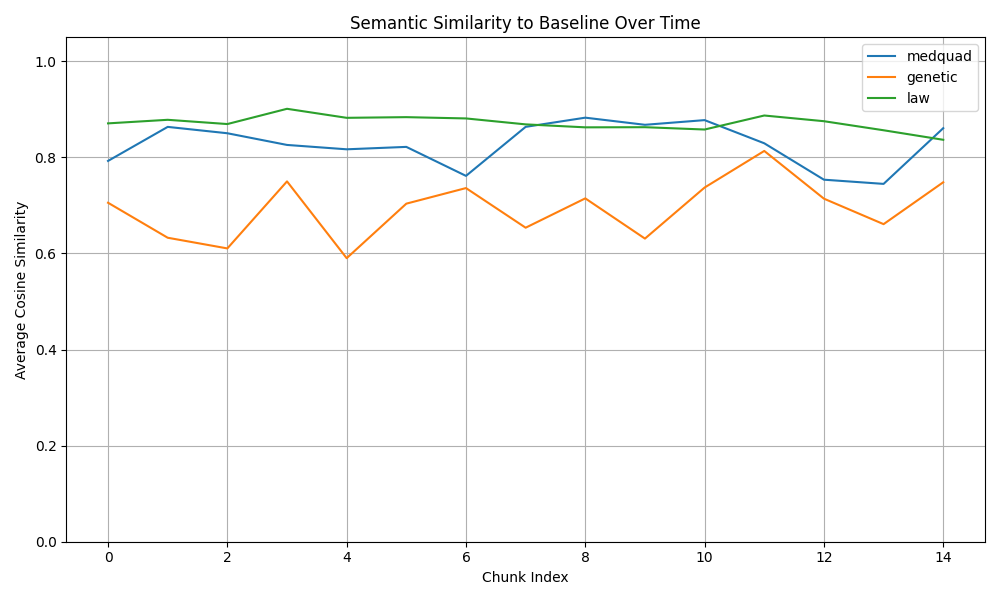}
    \caption{Cosine similarity to baseline answers over time}
    \label{fig:cosine}
\end{figure}
Figure 2 shows average semantic similarity between current and baseline answers, measured using cosine similarity. A higher score indicates greater preservation of baseline semantics. The Law domain retains high similarity throughout, while MedQuAD fluctuates mildly but remains above 0.8, suggesting stable retention. In contrast, the Genetic domain undergoes pronounced drift in the early chunks, indicating semantic divergence from original knowledge, with only partial recovery observed after replay. This supports the perplexity findings and reflects semantic degradation due to new domain interference.
\begin{figure}[H]
    \centering
    \includegraphics[width=0.7\linewidth]{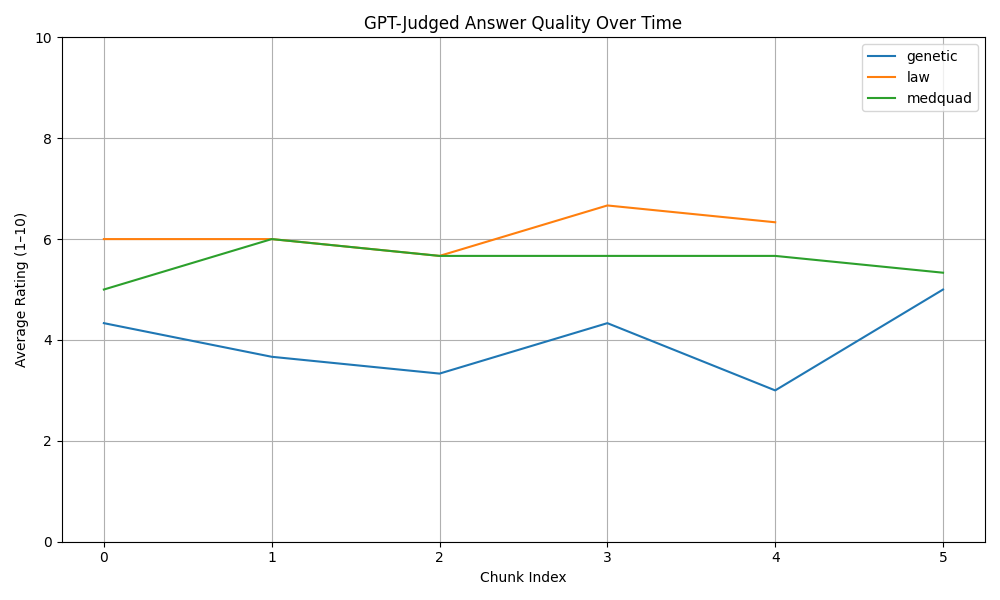}
    \caption{GPT-4 answer ratings across streaming chunks}
    \label{fig:gpt_scores}
\end{figure}
Figure 3 presents average human-like answer quality ratings (1–10) from GPT-4 for each domain over time. The Law domain consistently scores highest, reinforcing its stability across metrics. MedQuAD shows minor variance with generally favorable evaluations, indicating strong baseline retention. The Genetic domain again stands out for its early decline, with scores dipping below 4 in mid-streaming but recovering toward the end. These qualitative assessments align with perplexity and similarity metrics and highlight replay's role in regaining communicative and factual quality.

\section{Discussion}

Our results provide compelling evidence that real-time domain adaptation using LoRA and a lightweight replay buffer can mitigate catastrophic forgetting in large language models under constrained conditions. Across all three domains, we observed domain-sensitive variations in both forgetting and recovery. The MedQuAD domain, for instance, exhibited moderate perplexity increase and semantic drift, but maintained relatively high GPT-4 rating scores, suggesting resilience in domains with moderately complex factual content.

The Law domain presented an interesting case: despite domain shifts, it retained low perplexity, high semantic similarity, and consistently high human-evaluated scores. This could be attributed to either intrinsic regularity in the structure of legal language or the model's pretraining exposure to legal datasets. Conversely, the Genetic domain was highly volatile, suffering sharp spikes in perplexity and semantic deterioration, coupled with low GPT-4 ratings, but also demonstrated the most visible recovery post-replay, indicating potential for gradual re-learning with sufficient reinforcement.

Semantic similarity, while helpful, provided a more mechanical measure of knowledge retention and was sensitive to shifts in answer phrasing rather than factual correctness \citep{bogolin2023semantic}. In contrast, the GPT-based evaluation revealed qualitative nuances that the other metrics could not capture, such as shifts in explanatory depth, reasoning completeness, or nuanced medical detail. Taken together, these multi-faceted metrics provided a robust lens for interpreting real-time model behavior.

Another noteworthy observation was the system's ability to prevent complete degradation even without continuous replay \citep{unfoldai2024forgetting}. This resilience implies that LoRA-based adapters have an inherent capacity to preserve structural or factual patterns from earlier domains \citep{nightfall2024forgetting}. Yet, improvement beyond the baseline was sparse, indicating a need for additional mechanisms if true domain specialization is the goal , such as adapter stacking, per-domain optimization, or meta-learning initialization.

Finally, our approach demonstrates the feasibility of deploying dynamic, self-stabilizing LLMs in practical settings , without massive retraining or sophisticated infrastructure. This positions our work not just as a continual learning study, but as a foundation for edge-deployed, adaptable AI agents that serve users across evolving knowledge contexts.

\section{Limitations}
Despite these promising findings, our work has several limitations. First, our replay mechanism was fixed in size and did not prioritize samples based on utility or informativeness. More advanced strategies such as reservoir sampling, adaptive replay, or domain-aware rehearsal could yield better results.

Second, our evaluation was limited to a small set of prompts per domain and did not cover edge cases or out-of-distribution queries. This limits the generalizability of our conclusions to broader task diversity within each domain.

Third, the semantic similarity metric does not perfectly correlate with human judgment \citep{aguru2024similarity}. A model might generate more informative or accurate answers that differ semantically from the original, resulting in a lower similarity score despite qualitative improvement.

Lastly, we did not experiment with adapter merging, task-specific LoRA routing, or multi-adapter fusion, techniques that could allow models to preserve multiple domain competencies without interference. Future work could explore such modularity in dynamic inference settings.

Overall, this study serves as a foundation for building real-time adaptive language models that are both lightweight and robust, while also identifying clear areas for extension and improvement.

\section{Conclusion}

This work explored the viability of real-time domain adaptation in large language models using LoRA-based fine-tuning and a lightweight replay strategy. Across three distinct knowledge domains, medical, genetic, and legal, we simulated streaming adaptation conditions and tracked performance using perplexity, semantic similarity, and GPT-based answer quality ratings. Our findings underscore several key contributions:

\begin{itemize}
    \item \textbf{Resilience to forgetting:} Even under streaming conditions with limited replay, LoRA-equipped models maintained reasonable performance and avoided total collapse of prior knowledge, particularly in stable domains like Law.
    \item \textbf{Domain-specific drift and recovery:} The Genetic domain demonstrated the most severe forgetting and the clearest signal of recovery following replay, highlighting the importance of memory-based reinforcement in volatile domains.
    \item \textbf{Value of multi-metric evaluation:} Combining human-aligned ratings with classical metrics provided a nuanced understanding of how models evolve during adaptation, not merely drifting semantically, but also gaining or losing fluency, completeness, and specificity.
    \item \textbf{Operational simplicity:} Our approach relies on minimal infrastructure and can be deployed in environments with modest computational resources, opening doors for practical continual learning use cases in production.
\end{itemize}

While we did not observe significant performance gains beyond the baseline, our study validates that adaptive retention and domain-sensitive stability are achievable without catastrophic forgetting. These findings lay the groundwork for further research into modular, lifelong learning LLMs, especially in areas requiring frequent updates, personalized behavior, or embedded intelligence.

Future directions include dynamic replay prioritization, domain-specific adapter routing, and zero-shot recovery in out-of-domain tasks \citep{raschka2023tips}. By further improving adaptability without incurring high retraining costs, LLMs can become truly responsive agents capable of lifelong, context-aware learning.

\bibliographystyle{iclr2025_conference}

\end{document}